\documentclass{article}
\usepackage{PRIMEarxiv}
\usepackage[utf8]{inputenc} 
\usepackage[T1]{fontenc}    
\usepackage{hyperref}       
\usepackage{url}            
\usepackage{booktabs}       
\usepackage{amsfonts}       
\usepackage{nicefrac}       
\usepackage{microtype}      
\usepackage{lipsum}
\usepackage{fancyhdr}       
\usepackage{graphicx}       
\usepackage[title]{appendix}
\usepackage[square,numbers]{natbib}
\usepackage{float}
\usepackage{amsthm}
\usepackage{caption}
\usepackage[framemethod=tikz]{mdframed}
\usepackage{etoolbox} 
\usepackage{orcidlink}
\usepackage{listings}
\usepackage{tabularx}
\usepackage{multirow}
\usepackage{longtable}
\usepackage{booktabs}
\usepackage{makecell}
\usepackage{xcolor}
\usepackage{amsmath} 
\graphicspath{{media/}} 

\theoremstyle{definition}

\newmdtheoremenv[
  hidealllines=true,
  leftline=true,
  innerleftmargin=10pt,
  innerrightmargin=10pt,
  skipabove=10pt,
  skipbelow=10pt,
]{prompt}{Prompt}

\title{Revisiting Northrop Frye's Four Myths Theory with Large Language Models}

\author{
  Edirlei Soares de Lima \orcidlink{0000-0002-2617-3394}\\
  Academy for AI, Games and Media \\
  Breda University of Applied Sciences \\
  Breda, The Netherlands\\
  \texttt{soaresdelima.e@buas.nl} \\
  \And
  Marco A. Casanova \orcidlink{0000-0003-0765-9636}\\
  Department of Informatics \\
  PUC-Rio \\
  Rio de Janeiro, Brazil\\
  \texttt{casanova@inf.puc-rio.br} \\
  \And
  Antonio L. Furtado \orcidlink{0000-0003-3710-624X}\\
  Department of Informatics \\
  PUC-Rio \\
  Rio de Janeiro, Brazil\\
  \texttt{furtado@inf.puc-rio.br} \\
}

\begin{document}
\maketitle

\begin{abstract}
Northrop Frye's theory of four fundamental narrative genres (comedy, romance, tragedy, satire) has profoundly influenced literary criticism, yet computational approaches to his framework have focused primarily on narrative patterns rather than character functions. In this paper, we present a new character function framework that complements pattern-based analysis by examining how archetypal roles manifest differently across Frye's genres. Drawing on Jungian archetype theory, we derive four universal character functions (protagonist, mentor, antagonist, companion) by mapping them to Jung's psychic structure components. These functions are then specialized into sixteen genre-specific roles based on prototypical works. To validate this framework, we conducted a multi-model study using six state-of-the-art Large Language Models (LLMs) to evaluate character-role correspondences across 40 narrative works. The validation employed both positive samples (160 valid correspondences) and negative samples (30 invalid correspondences) to evaluate whether models both recognize valid correspondences and reject invalid ones. LLMs achieved substantial performance (mean balanced accuracy of 82.5\%) with strong inter-model agreement (Fleiss' $\kappa$ = 0.600), demonstrating that the proposed correspondences capture systematic structural patterns. Performance varied by genre (ranging from 72.7\% to 89.9\%) and role (52.5\% to 99.2\%), with qualitative analysis revealing that variations reflect genuine narrative properties, including functional distribution in romance and deliberate archetypal subversion in satire. This character-based approach demonstrates the potential of LLM-supported methods for computational narratology and provides a foundation for future development of narrative generation methods and interactive storytelling applications.
\end{abstract}


\section{Introduction}

As part of our long-term project in computational narratology and interactive story composition, we return to Northrop Frye's seminal work \cite{Frye2020}, focusing on his book's third essay entitled ``Archetypal Criticism: Theory of Myths''. Our objective is to model the \textit{character functions} underlying the four fundamental genres that Frye associated with the seasons' cyclic succession. We propose to complement our previous approach to Frye's theory, formulated in terms of \textit{narrative patterns} \cite{Lima2024Multigenre}, by developing a character-based approach grounded in Carl Gustav Jung's analytical psychology \cite{Jacobi2013}.

Similarly to the seasonal succession -- starting with merry spring, followed by the plenitude of summer, then by the falling decline of autumn, and ending in winter's desolation -- Frye's fundamental genres (\textit{comedy}, \textit{romance}, \textit{tragedy}, \textit{satire}) tell stories whose protagonists exhibit different degrees of power to influence action, combined with growing or decreasing happiness. The power of this seasonal metaphor extends beyond literary analysis: cyclic history researchers such as Oswald Spengler and Arnold Toynbee applied similar frameworks to civilizational rise and decline \cite{Sorokin1950}, demonstrating how recurring structural patterns can be identified across different domains through archetypal analysis. Just as these historians sought to identify the characteristic features distinguishing each historical phase, our character function approach seeks to identify the distinctive ways archetypal roles manifest across Frye's fundamental genres. This requires moving beyond recognizing that genres differ to specifying \textit{how} character functions operate differently within each genre's structural logic.

To achieve this specification, we identified four character functions through correspondence with Jungian psychic structure: \textit{protagonist}, \textit{mentor}, \textit{antagonist}, and \textit{companion}. We then specialized these functions by devising genre-specific \textit{role names} that express how each function operates within a particular genre's conventions. For each genre, we selected one prototypical work, identified characters embodying each function, and formulated role names expressing the specialized actions inherent to each function in that narrative context. 

To evaluate whether these character-role correspondences capture systematic patterns beyond the four prototypical works, we conducted a multi-model validation study using six state-of-the-art Large Language Models (LLMs) as analytical instruments. The validation employed both positive samples (valid character-role correspondences across 40 works) and negative samples (systematically introduced invalid correspondences) to assess whether the framework exhibits both descriptive adequacy (recognizing valid implementations across diverse narratives) and discriminative power (rejecting functionally incorrect assignments). Analysis of model reasoning patterns in error cases provides additional insight into the interpretive challenges inherent in character function identification and the boundaries of the functional role framework.

The paper is organized as follows. Section 2 reviews Frye's genre characterization and presents our character functions proposal grounded in Jungian archetypes. Section 3 describes the validation methodology and presents results examining overall performance, inter-model agreement, and performance patterns across genres and roles, followed by discussion of the findings. Section 4 provides concluding remarks.

\section{The Character Functions Approach}

In the third essay of his book \textit{Anatomy of Criticism} \cite{Frye2020}, entitled ``Archetypal Criticism: Theory of Myths'', Northrop Frye associates each of four fundamental literary genres with a season of the year. The seasonal succession, in turn, corresponds to the progression of human life from youth to old age. Frye's schema, together with exemplary works for each genre \cite{Moliere2013,Valmiki2022,Shakespeare2022,Orwell2021}, is presented in Table~\ref{tab:fundamental_genres}.

\begin{table}[h]
\centering
\caption{Frye's four fundamental genres with their seasonal and life-stage correspondences, and examples of prototypical works.}
\label{tab:fundamental_genres}
\begin{tabular}{llll}
\hline
Season & Age & Genre & Instance \\
\hline
Spring & adolescence & comedy & \textit{Le Bourgeois Gentilhomme} \\
Summer & youth & romance & \textit{Ramayana} \\
Autumn & maturity & tragedy & \textit{Macbeth} \\
Winter & senescence & satire & \textit{1984} \\
\hline
\end{tabular}
\end{table}

Comedies typically concern the efforts of an inexperienced young man to conquer a beloved damsel. A pompous older figure, known as an \textit{alazon} in Greek comedy, attempts to block his pursuit, while another figure, remarkable for creative practical talent and known as an \textit{eiron}, comes to his aid (Figaro, the resourceful servant, serving as a well-known example). These stories follow the spring phase of renewal and integration, leading to a happy ending in which the young protagonist overcomes old-fashioned privileges and social prejudices.

Romances, in the original epic sense, involve a quest whose objective is to counteract a villainy or obtain something marvelous. A hero is called to undertake the quest, but may require the guidance or magical instruments that a wise sage provides. The hero's inspiration and final reward is often the love of a highly virtuous woman.

Tragedies are the sequel of a nefarious act (\textit{hamartia}) that may be either mysteriously dictated by destiny or due to arrogant pride (\textit{hybris}) deserving dire punishment (\textit{nemesis}). Previous well-being is interrupted by the recognition of the act (\textit{anagnorisis}) followed by reversal of fortune (\textit{peripeteia}), as interpreted by oracular manifestation. The broken world order is painfully restored at the end.

Satires denounce dystopian world order, in the extreme case being characterized by the ``disappearance of the heroic'' \cite{Frye2020}. Protesters and their followers are punished for merely trying to behave normally by obedient guardians acting in the name of a supreme authority that can never be questioned.

To identify what might be the decisive character functions, ideally common to all genres, we looked at what Jung claimed to be the parts of every individual's psyche \cite{Jacobi2013}, namely: \textit{ego}, \textit{persona}, \textit{shadow}, \textit{anima}. The first two belong to consciousness -- the individual is continuously aware of existence through the ego part, while the persona ``is a compromise between the individual and society based on that which one appears to be'' \cite{Jacobi2013}. The last two are originally unconscious, the shadow constituting an alter ego wherein is kept whatever is rejected as negative, while the anima represents -- for a male -- a commonly negated feminine side.  

\begin{figure}
    \centering
    \includegraphics[width=0.4\linewidth]{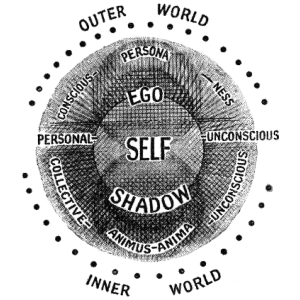}
    \caption{Diagram of Jung's psychic structure showing the four components (ego, persona, shadow, anima) and their integration through the individuation process \cite{Jacobi2013}.}
    \label{fig:psyche_diagram}
\end{figure}

A crucial resource of Jung's therapy is what he calls the \textit{individuation process}, whereby the level of consciousness is raised in order to allow the patient to take fuller advantage of originally underdeveloped potentials. The first step is to break the persona limitations. Quite often one's profession may exclusively determine what one believes to be. Next, shadow and anima, treated as archetypes, are brought to balanced consideration and their constructive traits are incorporated into consciousness. Another archetype then emerges as if by magic, the \textit{Wise Old Man}, which ``represents the cold and objective truth of nature ... in a variety of shapes ... well known from the world of primitives and from mythology'' \cite{Jacobi2013}. At this point the individuation process concludes, with the introduction of a new central personality focus, the \textit{self}, which from then on supplants the ego (as graphically suggested in Figure~\ref{fig:psyche_diagram}).

We are now in a position to introduce our proposed character functions, by simply attributing to separate characters what we have described as parts of the psyche of a single individual, as shown in Table~\ref{tab:jung_character_functions}. Note that in place of the persona -- the discarded masking part of the psyche (recalling that ``persona'' meant the mask used by actors to project the sound of their voices and identify their role to the audience of ancient theater performances) -- we have placed the revealing Wise Old Man archetype, where the ``old'' qualifier denotes superior wisdom not necessarily marked by advanced age.

\begin{table}[h]
\centering
\caption{Character functions derived from Jungian psychic structure.}
\label{tab:jung_character_functions}
\begin{tabular}{ll}
\hline
Part of psyche & Character function \\
\hline
Ego & protagonist \\
Persona ($\rightarrow$ Wise Old Man) & mentor \\
Shadow & antagonist \\
Anima & companion \\
\hline
\end{tabular}
\end{table}

In summary, the protagonist is recognized and oriented by the mentor, opposed by the antagonist, and inspired by the companion. The terms "mentor" and "shadow" also figure in Campbell and Vogler \cite{Campbell2008,Vogler2007}, both of whom, like Frye himself \cite{Frye2020}, acknowledge their debt to Jung. The "anima" archetype, described as a soul image whose "eternal feminine" figure shows Gretchen raising Faust's spirit to Heaven, is evoked in both Jungian studies \cite{Jacobi2013} and in Frye's book \cite{Frye2020}. Additionally, Jung-oriented researchers Emma Jung and Marie-Louise von Franz \cite{Jung1998} have argued that Grail romance narratives symbolically describe the entire individuation process, identifying the Grail as symbol of the Self.

We finally proceeded to specialize these character functions into typical \textit{character roles}, phrased to suggest how the functions might work differently in narratives pertaining to the different genres. To achieve this objective, we started from the prototype theory principle that classification is often achieved not through strict, necessary, and sufficient definitions but rather through similarity to a prototypical example \cite{Lakoff1990}. The titles shown earlier in the ``instances'' column of Table~\ref{tab:fundamental_genres}, namely \textit{Le Bourgeois Gentilhomme}, \textit{Ramayana}, \textit{Macbeth}, and \textit{1984}, were selected as prototypical based on their status as widely recognized exemplars of each genre.

The formulation of genre-specific role names required interpretive judgment about how character functions manifest in each prototype work. This interpretive dimension aligns with reader-response literary theory's recognition \cite{Iser1978} that literary analysis involves subjective engagement with texts rather than purely objective description. Based on analysis of the four prototypical works, we formulated role names expressing how each character function operates within genre-specific narrative contexts. The resulting role-character correspondences for the four prototypical works are listed below (in the format role:character):

\begin{itemize}
    \item \textit{Le Bourgeois Gentilhomme} (Comedy) -- lad in love:Cleonte; troubleshooter:Covielle; pompous blocker:Jourdain; witty damsel in love:Lucile
    \item \textit{Ramayana} (Romance) -- hero:Rama; donor:Visvamitra; villain:Ravana; faithful victim:Sita
    \item \textit{Macbeth} (Tragedy) -- ill-fated adventurer:Macbeth; soothsayer:three witches; order restorer:Macduff; ill-fated partner:Lady Macbeth
    \item \textit{1984} (Satire) -- nonconformist:Winston; inquisitor:O'Brien; dystopian idol:Big Brother; rebellion partner:Julia
\end{itemize}

Table~\ref{tab:genre_narrative_roles} presents these specialized role names organized by their corresponding character functions.

\begin{table}[h]
\centering
\caption{Specialized character roles for each of Frye's four fundamental genres.}
\label{tab:genre_narrative_roles}
\begin{tabular}{lllll}
\hline
Genre & Protagonist & Mentor & Antagonist & Companion \\
\hline
Comedy & lad in love & troubleshooter & pompous blocker & witty damsel in love \\
Romance & hero & donor & villain & faithful victim \\
Tragedy & ill-fated adventurer & soothsayer & order restorer & ill-fated partner \\
Satire & nonconformist & inquisitor & dystopian idol & rebellion partner \\
\hline
\end{tabular}
\end{table}

The resulting framework maps four universal character functions (protagonist, mentor, antagonist, companion) to sixteen genre-specific roles, as presented in Table~\ref{tab:genre_narrative_roles}. These specialized role names are designed to capture how archetypal functions manifest differently across Frye's four fundamental genres. The framework's validity depends on whether these prototype-derived correspondences generalize to other narratives within each genre. A theoretical consideration arises from Frye's recognition of ``intermediate phases'' consisting of narratives that combine features of more than one genre \cite{Frye2020}. Shakespeare's tragedies provide a well-known example, as we are told in \cite{Durant1961} how he ``learned the difficult art of intensifying tragedy with comic relief''. Genres should therefore be understood as overlapping clusters around prototypical examples rather than categories with strict boundaries, with individual works exhibiting varying degrees of alignment with genre conventions. The following section presents a systematic validation study that evaluates whether the proposed character-role correspondences capture recognizable patterns across diverse narratives beyond the four prototypical works.

\section{Experimental Validation}

To evaluate the validity of our character function framework beyond the four prototypical works from which it was derived, we conducted a multi-model validation study. Using six LLMs as analytical instruments, we assessed whether the proposed character-role correspondences capture systematic narrative patterns through evaluation on both positive samples (valid character-role correspondences across 40 works) and negative samples (30 invalid correspondences systematically introduced in the dataset). The experiment tests both the framework's capacity to describe character functions across diverse narratives and its discriminative power to reject functionally incorrect assignments.

\subsection{Validation Dataset}

\subsubsection{Sample Generation and Validation} \label{subsec:sample_generation}

To construct a dataset for validating the proposed character functions, we employed a semi-automated approach combining LLM-assisted identification with expert validation. We utilized GPT-5.2 (gpt-5.2-2025-12-11) to identify literary works and propose character-role correspondences, which were then reviewed by the authors to verify that: (1) the work was a recognized exemplar of the specified genre; and (2) the proposed character-role assignments reflected each character's primary structural function in the narrative. We acknowledge that this validation process involves interpretive judgment; however, the purpose of this dataset is not to establish definitive ``ground truth'' but rather to provide a consistent set of theoretically motivated correspondences against which to evaluate whether multiple independent LLMs recognize similar functional patterns. The subsequent multi-model evaluation thus tests whether our framework-based interpretations align with patterns recognizable to different models trained on diverse literary corpora.

For each of the four fundamental genres (comedy, romance, tragedy, and satire), we identified 10 exemplar works through an iterative process. We provided the model with the genre and its four corresponding functional roles from Table~\ref{tab:genre_narrative_roles}, instructing it to propose a work from that genre and identify which characters embodied each functional role. The model was explicitly directed to focus on structural character functions (what characters do in the story) rather than psychological complexity or thematic interpretation, aligning with our emphasis on plot-level functional analysis derived from structural narratology frameworks. The complete generation prompt is provided in Appendix~\ref{appendixA}. Works were identified individually to ensure independent selection without contextual influence from previously suggested examples. For each proposed work, we verified the two criteria mentioned above, repeating the process when proposed correspondences were ambiguous or incorrect.

This approach ensured dataset quality and theoretical coherence while exploring the model's domain knowledge to identify diverse works spanning different time periods, cultural contexts, and narrative media. The resulting dataset encompasses theatrical works (e.g., Shakespeare's \textit{Much Ado About Nothing}, Molière's \textit{Le Bourgeois Gentilhomme}), classic literature (e.g., Austen's \textit{Pride and Prejudice}, Orwell's \textit{1984}), ancient epics (e.g., Valmiki's \textit{Ramayana}, Sophocles' \textit{Oedipus Rex}), and modern film and television (e.g., \textit{The Hunger Games}, the 1967 TV series \textit{The Prisoner}). This range ensures that our functional role framework is evaluated across varied narrative implementations rather than being constrained to specific historical or cultural conventions.

The complete dataset comprises 40 works (10 per genre), yielding 160 character-role correspondences (40 works $\times$ 4 roles per work). The full list of works with their validated character-role assignments is provided in Appendix~\ref{appendixB}. We refer to this dataset as the \textit{positive samples}, as these correspondences represent theoretically valid functional assignments according to our framework.

\subsubsection{Negative Sample Construction} \label{subsec:negative_construction}

To evaluate whether LLMs can discriminate between valid and invalid character-role correspondences, we constructed a dataset of negative samples by introducing errors into a subset of the positive correspondences. Similar to the positive sample identification process, we employed GPT-5.2 (gpt-5.2-2025-12-11) to propose incorrect character-role mappings.

From the 40 works in the positive dataset, we randomly selected 20 works (5 per genre, representing 50\% of the positive samples) for negative sample construction. This sampling approach ensures balanced genre representation while maintaining a substantial evaluation set for assessing discriminative ability. For each selected work, we generated an alternative set of four character-role mappings where one or two roles were reassigned to functionally incorrect characters, depending on the error strategy employed, while the remaining roles retained their correct assignments from the positive samples. We provided the model with the correct character-role correspondences from the positive sample and instructed it to create incorrect mappings by applying one of two predefined error strategies. The complete negative sample generation prompt is provided in Appendix~\ref{appendixC}.

The two substitution strategies were designed to create plausible but functionally incorrect correspondences that would meaningfully challenge the models' understanding of character roles. These strategies reflect two common interpretive errors in character function analysis:

\begin{enumerate}
    \item \textbf{Primary-secondary character swap}: Replacing the primary bearer of a role with a secondary or supporting character who performs related but subordinate functions within the narrative. For example, in \textit{Hamlet}, substituting Rosencrantz for Ophelia as the ``ill-fated partner'' creates a correspondence where a minor courtier is incorrectly elevated over the character whose tragic fate is central to the protagonist's downfall.
    \item \textbf{Inter-role character swap}: Exchanging characters between different roles within the same story. For example, in \textit{V for Vendetta}, swapping V and Evey's functional assignments incorrectly positions Evey as the ``nonconformist'' (protagonist function) while demoting V to ``rebellion partner'' (companion function), inverting their actual structural primacy in the narrative.
\end{enumerate}

Similar to the positive sample curation, the authors reviewed each generated negative sample to ensure that: (1) the proposed incorrect correspondence represented a genuine functional error according to our framework; (2) the error was plausible enough to require analytical discrimination rather than being trivially detectable; and (3) the substituted character actually existed in the work and had some narrative connection to the functional domain being tested.

The final negative dataset comprises 20 works with modified correspondences (5 per genre), yielding 30 incorrect character-role assignments: 20 from inter-role swap cases (10 stories $\times$ 2 swapped roles) and 10 from primary-secondary character swap cases (10 stories $\times$ 1 substituted role). The full list of negative samples with their incorrect character-role assignments is provided in Appendix~\ref{appendixD}.

\subsection{Evaluation Protocol}

To ensure reliable validation of our proposed character-role correspondences, we employed a multi-LLM evaluation approach. Since recognizing character roles in narrative works is a complex analytical task requiring understanding of narrative structure and plot-level functions, we evaluated correspondences across multiple models with different architectures and training data. This approach provides diverse analytical perspectives beyond the authors' interpretations while avoiding reliance on the biases or limitations of any single model. Consistent recognition or rejection of character-role correspondences across architecturally diverse models strengthens confidence that our framework captures systematic structural patterns rather than model-specific artifacts.

\subsubsection{Model Selection}

We selected six state-of-the-art LLMs for the validation experiment. The evaluated LLMs encompassed both proprietary and open models from multiple providers, including Anthropic, DeepSeek, Google, OpenAI, Meta, and Alibaba, representing diverse architectural approaches and training methodologies. The complete list of models, including their providers and specific versions, is presented in Table~\ref{tab:evaluated_models}.

\begin{table}[htbp]
\centering
\caption{Overview of the LLMs used in the validation study.}
\label{tab:evaluated_models}
\small
\begin{tabular}{lll}
\toprule
\textbf{Model} & \textbf{Description} & \textbf{Version} \\ 
\midrule
Claude Opus 4.5 & \raggedright Flagship model for complex tasks by Anthropic. & claude-opus-4-5-20251101 \\[2pt]
DeepSeek R1 (671B) & \raggedright 671-billion-parameter reasoning model by DeepSeek. & deepseek-r1:671b \\[2pt]
Gemini 2.5 Pro & \raggedright Advanced multimodal model by Google. & gemini-2.5-pro \\[2pt]
GPT OSS (120B) & \raggedright 120-billion-parameter open model by OpenAI. & gpt-oss:120b \\[2pt]
Llama 4 (128B-A17B) & \raggedright 128-billion-parameter mixture model by Meta. & llama4:128x17b \\[2pt]
Qwen 3 (235B) & \raggedright 235-billion-parameter model by Alibaba. & qwen3:235b \\[2pt]
\bottomrule
\end{tabular}
\end{table}

Proprietary models were accessed via their respective APIs, while open-source models were executed locally using a self-hosted Ollama server.\footnote{\href{https://ollama.com/}{https://ollama.com/}} All models were queried using identical parameters to ensure uniform evaluation conditions: temperature was set to 0.0 to promote deterministic outputs and maximize reproducibility, while all other parameters retained their default values. 

Notably, we excluded GPT-5.2 from the validation analysis despite its use in positive sample identification. While the expert curation process ensured that accepted correspondences reflected real examples rather than GPT-5.2's specific patterns, we conservatively excluded this model from validation to eliminate any potential for circularity bias.

\subsubsection{Prompt Design}

The evaluation task required LLMs to analyze character-role correspondences and determine whether each assignment could be justified based on the character's functional role in the narrative. The prompt was designed to focus model attention on structural functional analysis rather than thematic or psychological character interpretation, with particular emphasis on identifying the primary bearer of each role rather than accepting any character who performs related actions.

LLMs were provided with the story title, genre classification, and four character-role correspondences to evaluate. For each correspondence, models were instructed to determine whether the character served as the primary bearer of that role's function in the narrative structure. Responses included a binary justification judgment and a brief reasoning explanation (2-3 sentences). This structured output format enabled both quantitative performance measurement through the binary judgments and qualitative analysis of model reasoning patterns through the explanations.

The complete evaluation prompt and system prompt, including the specific output format specification, are provided in Appendix~\ref{appendixE}.

\subsubsection{Data Collection and Processing}

For each of the 40 works in the positive dataset and 20 works in the negative dataset, we generated character-role evaluations from all six models, resulting in a total of 360 evaluation instances (240 from positive samples and 120 from negative samples). Each positive sample evaluation consisted of four individual character-role assessments (one per functional role), yielding 960 judgments across all models (40 works $\times$ 4 roles $\times$ 6 models). For negative samples, we evaluated only the altered character-role correspondences: in inter-role swap cases (10 stories), both swapped roles were evaluated (20 roles $\times$ 6 models = 120 judgments), while in primary-secondary character swap cases (10 stories), only the single substituted role was assessed (10 roles $\times$ 6 models = 60 judgments). This yielded 1,140 total role-specific judgments for analysis (960 + 120 + 60).

We automatically parsed model responses to extract the structured justification judgment for each character-role correspondence, along with the accompanying reasoning explanation. For positive samples, a response was scored as correct (True Positive) if the model judged the correspondence as justified, indicating recognition of a valid functional assignment. For negative samples, only the altered correspondences were evaluated: a correct response (True Negative) required the model to judge the correspondence as not justified, indicating successful discrimination of the functionally incorrect assignment. The unchanged correspondences in each negative sample were excluded from analysis to prevent repeated evaluation of valid correspondences already assessed in the positive samples.

\subsubsection{Evaluation Metrics}

To assess model performance in validating our functional role framework, we computed classification metrics that evaluate both the ability to recognize valid character-role correspondences and to reject invalid ones. We evaluated three primary metrics:

\begin{itemize}
\item \textbf{Recall}: The proportion of valid correspondences correctly identified, measuring the model's ability to recognize functional role fulfillment. Formally, $\text{Recall} = \frac{TP}{TP + FN}$, where $TP$ represents true positives and $FN$ represents false negatives.
\item \textbf{Specificity}: The proportion of invalid correspondences correctly rejected, measuring the model's ability to discriminate functionally incorrect assignments. Formally, $\text{Specificity} = \frac{TN}{TN + FP}$, where $TN$ represents true negatives and $FP$ represents false positives.
\item \textbf{Balanced Accuracy}: The mean of recall and specificity, providing an overall performance measure that treats both recognition and discrimination capabilities equally. Formally, $\text{Balanced Accuracy} = \frac{Recall + Specificity}{2}$.
\end{itemize}

Balanced Accuracy serves as our primary performance indicator, as it captures both essential aspects of framework validation: recognizing valid correspondences and rejecting invalid ones. Both false acceptances (incorrectly accepting invalid correspondences) and false rejections (incorrectly rejecting valid correspondences) equally undermine confidence in the theoretical framework, making it important to weight recall and specificity equally regardless of sample sizes. Note that random guessing would yield a balanced accuracy of 50\%, providing a baseline for interpretation.

Performance was analyzed at three levels: (1) overall performance across all correspondences; (2) genre-specific performance to identify whether certain narrative structures present greater validation challenges; and (3) role-specific performance to determine whether particular functional roles prove more difficult to validate. Additionally, we conducted qualitative analysis of model reasoning patterns to understand the factors contributing to incorrect classifications.

\subsection{Results and Discussion}

\subsubsection{Overall Performance}

Table~\ref{tab:overall_performance} presents the overall validation performance across all six evaluated models. All models demonstrated strong discriminative ability, with balanced accuracy ranging from 79.9\% to 85.3\%, substantially exceeding the random baseline of 50\% by at least 29 percentage points.

\begin{table}[htbp]
\centering
\caption{Overall performance metrics for all evaluated models. Each model evaluated 160 positive correspondences and 30 negative correspondences (190 total per model, 1,140 across all models).}
\label{tab:overall_performance}
\small
\begin{tabular}{lcccccccc}
\toprule
\textbf{Model} & \textbf{Recall (\%)} & \textbf{Specificity (\%)} & \textbf{Balanced Accuracy (\%)} & \textbf{TP} & \textbf{FN} & \textbf{TN} & \textbf{FP} \\
\midrule
Claude Opus 4.5 & 80.6 & 90.0 & \textbf{85.3} & 129 & 31 & 27 & 3 \\
Qwen 3 (235B) & 76.9 & 93.3 & 85.1 & 123 & 37 & 28 & 2 \\
Gemini 2.5 Pro & 76.2 & 90.0 & 83.1 & 122 & 38 & 27 & 3 \\
GPT OSS (120B) & 71.9 & 90.0 & 80.9 & 115 & 45 & 27 & 3 \\
Llama 4 (128B-A17B) & 78.1 & 83.3 & 80.7 & 125 & 35 & 25 & 5 \\
DeepSeek R1 (671B) & 73.1 & 86.7 & 79.9 & 117 & 43 & 26 & 4 \\
\midrule
\textbf{Mean $\pm$ SD} & \textbf{76.1 $\pm$ 3.2} & \textbf{86.5 $\pm$ 3.6} & \textbf{81.4 $\pm$ 2.0} & & & & \\
\bottomrule
\end{tabular}
\end{table}

Claude Opus 4.5 achieved the highest overall performance (balanced accuracy of 85.3\%), demonstrating both strong recall (80.6\%) and specificity (90.0\%). This represents a near-optimal balance between recognizing valid correspondences and rejecting invalid ones. Qwen 3 (235B) followed closely with 85.1\% balanced accuracy, while Gemini 2.5 Pro achieved 83.1\%. The remaining models ranged from 79.9\% to 80.9\%.

An interesting pattern emerges when examining the recall-specificity trade-off across models. Most models exhibited higher specificity than recall, correctly rejecting invalid assignments (mean specificity of 88.9\%) more often than correctly identifying all valid ones (mean recall of 76.1\%). This pattern is most pronounced in Qwen 3 (235B), which achieved 93.3\% specificity while maintaining 76.9\% recall, and GPT OSS (120B) with 90.0\% specificity and 71.9\% recall. In contrast, Claude Opus 4.5 and Llama 4 maintained more balanced profiles, with recall rates closer to their specificity rates.

Performance remained consistent across the six models, with a standard deviation in balanced accuracy of 2.1\%. This consistency across architecturally diverse models suggests that evaluation patterns may reflect systematic properties of the narrative data. To examine this hypothesis more rigorously, we next analyze the extent to which models agree on individual character-role assessments.

\subsubsection{Model Agreement and Consensus}

Table~\ref{tab:agreement_summary} presents inter-model agreement metrics across all 190 character-role evaluations.

\begin{table}[htbp]
\centering
\caption{Summary of inter-model agreement metrics.}
\label{tab:agreement_summary}
\small
\begin{tabular}{lc}
\toprule
\textbf{Metric} & \textbf{Value} \\
\midrule
Mean pairwise agreement & 82.0\% $\pm$ 3.4\% \\
Pairwise agreement range & 76.3\% - 87.4\% \\
Unanimous agreement (6/6) & 113/190 (59.5\%) \\
Majority agreement (4+/6) & 159/190 (83.7\%) \\
Split decisions (3/3) & 11/190 (5.8\%) \\
Fleiss' Kappa ($\kappa$) & 0.600 \\
\bottomrule
\end{tabular}
\end{table}

Models demonstrated strong agreement across all metrics. Pairwise agreement averaged 82.0\% (SD = 3.4\%), with all 15 model pairs exceeding 75\% agreement (see the complete table in Appendix~\ref{appendixF}). Examining consensus across all six models simultaneously, 113 judgments (59.5\%) achieved unanimous agreement, with an additional 46 judgments (24.2\%) reaching majority consensus where 4-5 models agreed. Only 11 judgments (5.8\%) produced split decisions with models dividing evenly. Overall, 83.7\% of judgments showed clear majority agreement (4 or more models concurring).

To quantify inter-rater reliability, we calculated Fleiss' Kappa, which adjusts for chance agreement when multiple raters evaluate categorical data. The analysis yielded $\kappa = 0.600$, indicating substantial inter-rater reliability according to standard interpretation guidelines \cite{Landis1977}. This level of agreement substantially exceeds what would be expected by random chance (expected: 55.0\%, observed: 82.0\%).

\subsubsection{Performance by Genre}

The substantial inter-model agreement establishes that models converge on similar character-role assessments. Therefore, we aggregate evaluations across all six models to examine whether performance varies systematically across Frye's four fundamental genres. Table~\ref{tab:performance_by_genre} presents the aggregated metrics for each genre.

\begin{table}[htbp]
\centering
\caption{Performance metrics by genre, aggregated across all six models. Each genre includes 240 positive judgments (10 works $\times$ 4 roles $\times$ 6 models) and 42-48 negative judgments depending on error distribution.}
\label{tab:performance_by_genre}
\small
\begin{tabular}{lcccccc}
\toprule
\textbf{Genre} & \textbf{Recall (\%)} & \textbf{Specificity (\%)} & \textbf{Balanced Accuracy (\%)} & \textbf{Positive} & \textbf{Negative} & \textbf{Total} \\
 \midrule
Tragedy & 82.1 & 97.6 & \textbf{89.9} & 240 & 42 & 282 \\
Comedy & 79.6 & 90.5 & 85.0 & 240 & 42 & 282 \\
Romance & 67.9 & 89.6 & 78.8 & 240 & 48 & 288 \\
Satire & 75.0 & 79.2 & 77.1 & 240 & 48 & 288 \\
\bottomrule
\end{tabular}
\end{table}

Performance varied notably across genres, spanning 12.8 percentage points from 77.1\% (satire) to 89.9\% (tragedy). Tragedy and comedy, representing Frye's autumn and spring phases respectively, achieved the highest performance with both strong recall and high specificity, suggesting that character-role correspondences in these genres present well-defined functional patterns. Romance maintained high specificity (89.6\%) but exhibited the lowest recall (67.9\%), indicating that models were conservative in accepting correspondences -- correctly rejecting invalid assignments but missing some valid ones. This pattern suggests that while incorrect romance correspondences are readily identifiable, valid correspondences may involve more subtle or varied functional implementations. Satire stands out with relatively balanced recall (75.0\%) and specificity (79.2\%), but both metrics substantially lower than other genres, suggesting that character functions in satirical narratives may be more ambiguous or context-dependent, challenging models to discriminate between valid and invalid correspondences.

\subsubsection{Performance by Role}

The substantial inter-model agreement also enables examining performance patterns across the 16 specialized character roles derived from the four functional archetypes. Table~\ref{tab:performance_by_role} presents performance metrics aggregated across all six models for each role.

\begin{table}[htbp]
\centering
\caption{Performance metrics by role, aggregated across all six models. Each role includes 60 positive judgments (10 genre-specific works $\times$ 6 models). Negative judgments vary by role (6-18) based on error distribution in the negative sample construction.}
\label{tab:performance_by_role}
\small
\begin{tabular}{llccccc}
\toprule
\textbf{Role} & \textbf{Genre} & \textbf{Recall (\%)} & \textbf{Specificity (\%)} & \textbf{Balanced Accuracy (\%)} & \textbf{Positive} & \textbf{Negative} \\
\midrule
\multicolumn{7}{l}{\textit{Protagonist function}} \\
Nonconformist & Satire & 98.3 & 100.0 & \textbf{99.2} & 60 & 12 \\
Ill-fated adventurer & Tragedy & 93.3 & 100.0 & 96.7 & 60 & 6 \\
Lad in love & Comedy & 85.0 & 100.0 & 92.5 & 60 & 12 \\
Hero & Romance & 78.3 & 100.0 & 89.2 & 60 & 6 \\
\midrule
\multicolumn{7}{l}{\textit{Mentor function}} \\
Troubleshooter & Comedy & 85.0 & 83.3 & 84.2 & 60 & 12 \\
Donor & Romance & 61.7 & 100.0 & 80.8 & 60 & 12 \\
Soothsayer & Tragedy & 70.0 & 83.3 & 76.7 & 60 & 6 \\
Inquisitor & Satire & 78.3 & 72.2 & 75.3 & 60 & 18 \\
\midrule
\multicolumn{7}{l}{\textit{Antagonist function}} \\
Order restorer & Tragedy & 83.3 & 100.0 & 91.7 & 60 & 12 \\
Dystopian idol & Satire & 76.7 & 100.0 & 88.3 & 60 & 6 \\
Villain & Romance & 75.0 & 100.0 & 87.5 & 60 & 12 \\
Pompous blocker & Comedy & 78.3 & 83.3 & 80.8 & 60 & 6 \\
\midrule
\multicolumn{7}{l}{\textit{Companion function}} \\
Ill-fated partner & Tragedy & 81.7 & 100.0 & 90.8 & 60 & 18 \\
Witty damsel in love & Comedy & 70.0 & 91.7 & 80.8 & 60 & 12 \\
Faithful victim & Romance & 56.7 & 72.2 & 64.4 & 60 & 18 \\
Rebellion partner & Satire & 46.7 & 58.3 & 52.5 & 60 & 12 \\
\bottomrule
\end{tabular}
\end{table}

Performance varied substantially across roles, with balanced accuracy ranging from 52.5\% to 99.2\%. Protagonist-function roles consistently achieved the highest performance across all genres, with all four exhibiting perfect specificity (100.0\%) and strong recall (78.3\% - 98.3\%). In contrast, companion roles exhibited the widest performance variation (52.5\% - 90.8\%), spanning 38.3 percentage points.

Eight roles achieved perfect specificity (100.0\%), all belonging to either protagonist or antagonist functions. However, several roles exhibited notable recall-specificity gaps, with ``donor'' (romance) showing the largest disparity at 38.3 percentage points (61.7\% recall vs. 100.0\% specificity). This pattern was also observed in ``villain'' (romance) and ``dystopian idol'' (satire), where models more consistently rejected invalid correspondences than identified all valid ones. The lowest-performing roles cluster within specific genre-function combinations: ``rebellion partner'' (satire, 52.5\%), ``faithful victim'' (romance, 64.4\%), and ``inquisitor'' (satire, 75.3\%). These three roles all belong to either companion or mentor functions in romance and satire genres, suggesting differences in performance across genre-function pairings.

\subsubsection{Discussion}

The validation results support the theoretical premise that Jungian archetypes, when specialized according to Frye's genre distinctions, provide a promising framework for analyzing character functions in narrative works. LLMs achieved substantial performance (mean balanced accuracy of 82.5\%) with strong inter-model agreement (Fleiss' $\kappa$ = 0.600 and mean pairwise agreement of 82.0\%). The consistency of this performance across six architecturally diverse models suggests that the observed patterns reflect structural properties of the narrative data rather than model-specific biases. Moreover, the character-role correspondences derived from four prototypical works generalized effectively across forty diverse narratives spanning different time periods, cultural contexts, and media, suggesting that the framework captures fundamental rather than idiosyncratic patterns of character function.

The genre-level performance differences align with Frye's characterization of these narrative forms. Tragedy and comedy follow relatively structured narrative arcs with clearly defined character functions: tragedy progresses from prosperity through hamartia to catastrophe and the restoration of moral order, while comedy moves from social obstruction to integration and renewal \cite{Frye2020}. In tragedy, the order restorer serves a clearly delineated function -- restoring the moral order disrupted by the protagonist's transgression -- while in comedy, the troubleshooter operates within well-established conventions to overcome the blocking figure's obstruction. These established structural patterns create consistent expectations for how characters fulfill their functions, facilitating reliable character-role identification.

Romance and satire exhibit structural properties that introduce ambiguity in character-role identification. Romance's quest structure exhibits the variability that Propp observed in folktale morphology, where not all narrative functions manifest in every story and helper figures take diverse forms \cite{Propp1968}. A hero may receive aid from a sage who provides counsel, from magical objects without a clear donor figure, or may overcome obstacles through inherent prowess without distinct antagonistic characters. When functions can be distributed across multiple characters, absent entirely, or fulfilled through non-character elements, determining which character serves as the primary bearer of a role becomes ambiguous. Qualitative examination of model reasoning patterns for the ``donor'' role (which achieved perfect specificity but only 61.7\% recall) provides concrete evidence for the functional distribution characteristic of romance narratives. When evaluating Mrs. Fairfax as donor in \textit{Jane Eyre}, multiple models identified alternative donors serving different quest phases: Claude Opus 4.5 noted ``Miss Temple (who nurtures Jane's education)'' and ``Jane's uncle who leaves her the inheritance grants her independence'', while Qwen 3 emphasized that ``the primary donor role belongs to Jane's uncle John Eyre, who grants her financial independence''. Similarly, for Visvamitra in the \textit{Ramayana}, Gemini 2.5 Pro observed he provides ``divine weapons'' but is ``preliminary to the main plot'', with ``more central donors appear[ing] later, such as Sugriva who provides the army... and Vibhishana who provides the critical intelligence''. When the donor function is distributed across multiple characters and quest stages, determining which character serves as the primary bearer of the role becomes inherently ambiguous.

Satire exhibits a different form of ambiguity: Frye characterizes it as marked by the ``disappearance of the heroic'' \cite{Frye2020}, where protagonists lack the agency of traditional heroes and archetypal distinctions deliberately blur. Characters often fulfill functions in subverted or merged ways: a figure labeled ``inquisitor'' (our mentor-function role for satire) may simultaneously guide and manipulate the protagonist, as O'Brien does to Winston in \textit{1984}. This functional ambiguity creates legitimate disagreement about whether a character truly fulfills a role: is O'Brien primarily an inquisitor who reveals truth, or has the mentor function collapsed entirely into the antagonistic dystopian system? When archetypal functions are deliberately subverted or merged, even human readers might disagree on character-role assignments, making lower model performance an expected reflection of the genre's inherent interpretive ambiguity. Examination of model reasoning for the ``inquisitor'' role confirms this interpretation. When evaluating Effie Trinket as inquisitor in \textit{The Hunger Games}, multiple LLMs reasoned that while she enforces Capitol norms, the inquisitorial function more properly belongs to other entities: Claude Opus 4.5 identified ``characters who question or probe the protagonists' motives'', DeepSeek R1 noted ``the primary inquisitor function belongs to antagonists like Seneca Crane (Gamemaker surveillance) and President Snow himself'', and Gemini 2.5 Pro observed that ``the inquisitorial function... is performed by Capitol officials such as President Snow and the Gamemakers''. For Clevinger in \textit{Catch-22}, Gemini 2.5 Pro noted he is ``the victim of an inquisition, not the inquisitor himself'', while GPT OSS observed he is ``more a victim of bureaucratic interrogation than the agent who conducts it''. LLMs disagreed not because they failed to understand character functions, but because satire's deliberate subversion creates legitimate ambiguity about which character (if any single character) primarily embodies a given function.

The role-level patterns reveal that performance differences stem from both narrative structure and framework design choices. The consistently high performance of protagonist and antagonist roles across genres likely reflects their structural necessity: protagonists drive narrative action while antagonists create conflict, making these functions plot-central and relatively invariant across different stories. Companion functions, conversely, operate more at the relational and thematic level. A companion's primary contribution often lies in emotional support, moral influence, or thematic counterpoint rather than plot mechanics, allowing greater variation in how and whether these functions manifest as distinct character roles. The clustering of low performance in romance and satire companion roles (faithful victim with 64.4\% and rebellion partner with 52.5\%) may therefore reflect genuine structural differences in how these genres employ companion functions rather than inconsistent role characterizations.

Performance differences also reflect our framework design choices in formulating role names. The specialized role names vary considerably in their semantic specificity and clarity. Terms such as ``hero'' and ``villain'' invoke well-established archetypal concepts with relatively stable meanings, while ``rebellion partner'' or ``witty damsel in love'' combine multiple qualifiers that narrow the concept but potentially introduce interpretive ambiguity. The ``donor'' role borrows directly from Propp's established terminology \cite{Propp1968}, carrying the weight of a defined narratological function, while ``dystopian idol'' represents a more novel construction. These differences in label clarity may contribute to performance variation: roles with precise, established labels may facilitate recognition, while roles requiring interpretation of multiple qualifiers or novel combinations may be more difficult to generalize in multiple narrative implementations. Examination of model reasoning for the ``rebellion partner'' role (52.5\% balanced accuracy, the lowest overall) illustrates this challenge. When evaluating Clarisse McClellan in \textit{Fahrenheit 451}, LLMs consistently reasoned that while she ``catalyzes Montag's transformation'' (Claude Opus 4.5, DeepSeek R1, Gemini 2.5 Pro), she ``disappears early and does not actively participate in his rebellion'' (DeepSeek R1), with Faber serving as the ``operational partner'' (DeepSeek R1) or ``actively collaborat[ing] with Montag, provid[ing] a strategic plan'' (Gemini 2.5 Pro). For Evey Hammond in \textit{V for Vendetta}, Gemini 2.5 Pro noted her ``primary function is not that of a partner but of a successor'', while Llama 4 observed she is ``more supportive and developmental'' rather than an equal collaborator. Models grappled with whether partnership requires sustained collaboration, ideological equality, and shared agency, all questions without clear and consistent answers. Notably, in \textit{Brave New World}, five of six models accepted Helmholtz Watson instead of the designated Bernard Marx, with reasoning emphasizing that Helmholtz ``joins John in both intellectual dissent and active, physical rebellion'' (Gemini 2.5 Pro) while Bernard acts from ``self-interest and cowardice'' (Claude Opus 4.5, Gemini 2.5 Pro), revealing that models evaluated partnership quality (authenticity of commitment and depth of collaboration) rather than simply matching character to role labels. This reflects an inherent trade-off in our approach: genre-tailored role names capture the nuances Frye identifies in how archetypes manifest differently across genres, but this specificity comes at the cost of reduced generality.

\section{Concluding Remarks}

This work establishes a character function framework that bridges Frye's literary genre theory with computational narrative analysis. By deriving four universal functions from Jungian archetypes and specializing them into sixteen genre-specific roles, we provide a formalized structure for representing how characters operate within different narrative genres. The validation demonstrates that these character-role correspondences capture systematic patterns across diverse works. LLMs successfully recognized valid correspondences while rejecting invalid ones, with performance variations reflecting genuine structural differences in how character functions manifest across genres. This framework offers computational narratology a practical resource for narrative analysis and generation, with the specialized roles providing explicit functional specifications that can guide character behavior in interactive storytelling systems, inform automated story analysis, or serve as constraints in narrative generation algorithms.

While the validation demonstrates the framework's viability, several considerations define its current scope and indicate possible directions for future research. Although our validation included narratives with both male and female protagonists, the theoretical foundation in Jung's male-centered archetype theory raises questions about gender-specific manifestations of character functions. Recent research has demonstrated that LLMs exhibit gender biases in narrative generation, often using protagonist gender as a heuristic for interpreting narrative structures \cite{VanBlerck2025}. In this context, Jung identifies different psychic structures for women (animus rather than anima, Great Mother rather than Wise Old Man), which could theoretically affect how archetypal functions map to character roles in female-centered narratives. Whether these theoretical differences manifest as systematic patterns in actual narrative analysis remains an empirical question for future investigation, potentially drawing on feminist revisions of archetypal theory such as Murdock's heroine's journey \cite{Murdock1990} or Pratt's work on archetypal patterns in women's fiction \cite{Pratt1981}. Additionally, the specialized role names reflect interpretive choices made through analysis of the four prototypical works. While the validation demonstrates that these prototype-derived roles generalize across diverse works within each genre, alternative prototype selections might yield different role formulations that prove equally valid. The observed trade-off between genre-specific precision and cross-genre generality suggests that role name specificity, while capturing important genre distinctions, may limit broader applicability.

Future work could also expand the functional scope to encompass additional character types that Frye identifies as intensifying genre-specific emotional effects. While our framework focuses on the four archetypal functions necessary to drive narrative action (protagonist, mentor, antagonist, companion), Frye notes that each genre employs supplementary characters who amplify its characteristic mood. In comedy, \textit{buffoons} and \textit{churls} complement the eiron (our \textit{troubleshooter}) and alazon (pompous blocker) to polarize the comic atmosphere \cite{Frye2020}. In romance, magical helpers beyond the primary donor perform extraordinary feats to render the hero's achievement more remarkable \cite{Propp1968}. Tragedy may feature denouncers or supernatural agents who intensify the emotions of fear and pity that Aristotle identified as essential to tragic catharsis \cite{Aristotle2001}, while satire often employs hate symbol or scapegoat figures who concentrate blame and deflect attention from systemic failures, intensifying the genre's critical mood. Investigating whether these mood-intensifying characters follow systematic patterns across genres, and how they interact with the core functional roles, represents a promising direction for enriching the character function framework while maintaining its theoretical grounding in archetypal structures.

The proposed framework provides computational narratology with a theoretically grounded resource for character-based narrative generation and analysis. The sixteen specialized roles offer explicit functional specifications that can support the development of computational narrative generation methods and interactive storytelling applications. Previous work has demonstrated that character-based interactive storytelling can be achieved through multi-agent planning systems, where autonomous character agents interact within plot management frameworks to generate coherent narratives in highly interactive game environments \cite{Lima2023,Lima2022}. The framework proposed in this work can extend these approaches by providing genre-specific functional roles that LLM-based character agents could embody. This character function approach complements our recent work on LLM-based narrative generation methods that employ semiotic reconstruction \cite{Lima2025,Lima2024ChatGeppetto} and narrative pattern guidance \cite{Lima2024Pattern,Lima2025Player}, enabling systems to operate at both the pattern level (plot structure) and the character level (functional roles). The integration of character functions with pattern-based methods advances toward more sophisticated computational narrative systems that maintain both structural coherence and genre-appropriate character behavior.

This work demonstrates that computational methods can engage productively with established literary theory, not merely applying humanistic concepts to computational tasks but using computational validation to illuminate the theoretical constructs themselves. The convergence of multiple independent LLMs on similar character-role assessments, combined with the interpretive insights revealed through error analysis, suggests that LLM-supported methods offer computational narratology valuable instruments for testing, refining, and extending narratological frameworks. As these methods mature, the integration of theoretically grounded frameworks with computational validation may bridge the longstanding gap between humanistic literary scholarship and computational approaches to narrative generation.

\bibliographystyle{abbrvnat}
\bibliography{references}

@book{Campbell2008,
  author    = {Campbell, Joseph},
  title     = {The Hero with a Thousand Faces},
  publisher = {New World Library},
  year      = {2008},
  address   = {Novato, California}
}

@book{Frye2020,
  author    = {Frye, Northrop},
  title     = {Anatomy of Criticism: Four Essays},
  publisher = {Princeton University Press},
  year      = {2020},
  address   = {Princeton, New Jersey}
}

@book{Jacobi2013,
  author    = {Jacobi, Jolande},
  title     = {Psychology of C. G. Jung},
  publisher = {Routledge},
  year      = {2013},
  address   = {London}
}

@book{Jung1998,
  author    = {Jung, Emma and von Franz, Marie-Louise},
  title     = {The Grail Legend},
  publisher = {Princeton University Press},
  year      = {1998},
  address   = {Princeton, New Jersey}
}

@misc{Lima2024Multigenre,
  author       = {de Lima, Edirlei Soares and Neggers, Margot M. E. and Furtado, Antonio L.},
  title        = {Multigenre AI-powered Story Composition},
  year         = {2024},
  eprint       = {2405.06685},
  archiveprefix = {arXiv},
  primaryclass = {cs.CL},
  doi          = {10.48550/arXiv.2405.06685}
}

@book{Moliere2013,
  author    = {Molière},
  title     = {Le Bourgeois Gentilhomme},
  publisher = {Folio},
  year      = {2013},
  address   = {Paris}
}

@book{Orwell2021,
  author    = {Orwell, George},
  title     = {Nineteen Eighty-Four},
  publisher = {Wordsworth Editions},
  year      = {2021},
  address   = {Ware, Hertfordshire}
}

@book{Propp1968,
  author    = {Propp, Vladimir},
  title     = {Morphology of the Folktale},
  publisher = {University of Texas Press},
  year      = {1968},
  address   = {Austin, Texas}
}

@book{Shakespeare2022,
  author     = {Shakespeare, William},
  title      = {Macbeth},
  publisher  = {SeaWolf Press},
  year       = {2022}
}

@book{Sorokin1950,
  author    = {Sorokin, Pitirim A.},
  title     = {Social Philosophies of an Age of Crisis},
  publisher = {Beacon Press},
  year      = {1950},
  address   = {Boston, Massachusetts}
}

@book{Valmiki2022,
  author    = {Vālmīki},
  title     = {The Rāmāyaṇa of Vālmīki: The Complete English Translation},
  publisher = {Princeton University Press},
  year      = {2022},
  address   = {Princeton, New Jersey},
  series    = {Princeton Library of Asian Translations},
  translator = {Goldman, Robert P. and Goldman, Sally J. Sutherland and Lefeber, Rosalind and Pollock, Sheldon I. and van Nooten, Barend A.},
}

@book{Vogler2007,
  author    = {Vogler, Christopher},
  title     = {The Writer's Journey: Mythic Structure for Writers},
  publisher = {Michael Wiese Productions},
  year      = {2007},
  address   = {Studio City, California} 
}

@book{Lakoff1990,
  author    = {Lakoff, George},
  title     = {Women, Fire, and Dangerous Things: What Categories Reveal about the Mind},
  publisher = {University of Chicago Press},
  year      = {1990},
  address   = {Chicago, Illinois}
}

@book{Iser1978,
  author     = {Iser, Wolfgang},
  title      = {The Act of Reading: A Theory of Aesthetic Response},
  publisher  = {Johns Hopkins University Press},
  year       = {1978},
  address    = {Baltimore, Maryland}
}

@book{Durant1961,
  author    = {Durant, Will and Durant, Ariel},
  title     = {The Age of Reason Begins: A History of European Civilization in the Period of Shakespeare, Bacon, Montaigne, Rembrandt, Galileo, and Descartes: 1558--1648},
  publisher = {Simon and Schuster},
  year      = {1961},
  address   = {New York},
}

@article{Landis1977,
  author  = {Landis, J. Richard and Koch, Gary G.},
  title   = {The Measurement of Observer Agreement for Categorical Data},
  journal = {Biometrics},
  year    = {1977},
  volume  = {33},
  number  = {1},
  pages   = {159--174},
  month   = {March},
  doi     = {10.2307/2529310}  
}

@book{Murdock1990,
  title     = {The Heroine's Journey: Woman's Quest for Wholeness},
  author    = {Murdock, Maureen},
  year      = {1990},
  publisher = {Shambhala},
  address   = {Boston, Massachusetts}
}

@book{Pratt1981,
  author    = {Pratt, Annis},
  title     = {Archetypal Patterns in Women's Fiction},
  publisher = {Indiana University Press},
  year      = {1981},
  address   = {Bloomington, Indiana}
}

@article{Lima2023,
  title = {Managing the plot structure of character-based interactive narratives in games},
  author = {de Lima, Edirlei Soares and Feijó, Bruno and Furtado, Antonio L.},
  journal = {Entertainment Computing},
  volume = {47},
  pages = {100590},
  year = {2023},
  issn = {1875-9521},
  doi = {10.1016/j.entcom.2023.100590}  
}

@inproceedings{Lima2022,
  author={de Lima, Edirlei Soares and Feijó, Bruno and Furtado, Antonio L.},
  booktitle={2022 21st Brazilian Symposium on Computer Games and Digital Entertainment (SBGames)}, 
  title={A Character-based Model for Interactive Storytelling in Games}, 
  year={2022},
  pages={1-6},  
  doi={10.1109/SBGAMES56371.2022.9961071}
}

@article{Lima2025,
  title = {An {AI}-powered approach to the semiotic reconstruction of narratives},
  journal = {Entertainment Computing},
  volume = {52},
  pages = {100810},
  year = {2025},
  doi = {10.1016/j.entcom.2024.100810},
  author = {de Lima, Edirlei Soares and Neggers, Margot M.E. and Feij\'{o}, Bruno and Casanova, Marco A. and Furtado, Antonio L.},
}

@inproceedings{Lima2024ChatGeppetto,
  author = {De Lima, Edirlei Soares and Feij\'{o}, Bruno and Cassanova, Marco A. and Furtado, Antonio L.},
  title = {{ChatGeppetto} - an {AI}-powered Storyteller},
  year = {2024},
  publisher = {ACM},
  address = {New York, NY, USA},
  doi = {10.1145/3631085.3631302},
  booktitle = {Proceedings of the 22nd Brazilian Symposium on Games and Digital Entertainment},
  pages = {28–37},  
  location = {Rio Grande (RS), Brazil}  
}

@inproceedings{Lima2024Pattern,
  author = {de Lima, Edirlei Soares and Neggers, Margot M. E.  and Casanova, Marco A. and Feijó, Bruno and Furtado, Antonio L.},
  editor={Figueroa, Pablo and Di Iorio, Angelo and Guzman del Rio, Daniel and Gonzalez Clua, Esteban Walter and Cuevas Rodriguez, Luis},
  title = {A Pattern-oriented {AI}-powered Approach to Story Composition},
  booktitle={Entertainment Computing -- ICEC 2024},
  publisher={Springer Cham},
  year = {2024},
  pages={1--16},
  doi = {10.1007/978-3-031-74353-5_10}
}

@InProceedings{Lima2025Player,
  author={de Lima, Edirlei Soares and Neggers, Margot M. E. and Casanova, Marco A. and Furtado, Antonio L.},
  editor={Marto, Anabela and Prada, Rui and Gouveia, Patr{\'i}cia and Espinosa, Ruth Contreras and Gon{\c{c}}alves, Alexandrino and Abrantes, Eduarda and Ribeiro, Roberto},
  title={From Images to Stories: Exploring Player-Driven Narratives in Games},
  booktitle={Videogame Sciences and Arts},
  year={2025},
  publisher={Springer Nature Switzerland},
  address={Cham},
  pages={228--242}
}

@book{Aristotle2001,
  editor     = {Murray, Penelope and Dorsch, T. S.},
  title      = {Classical Literary Criticism},
  publisher  = {Penguin Books},
  year       = {2001},
  address    = {London},
  pages      = {187},
  series     = {Penguin Classics},
  translator = {Murray, Penelope and Dorsch, T. S.},
}

@article{VanBlerck2025,
  title = {Unveiling gender bias in {LLM}-generated hero and heroine narratives},
  journal = {Entertainment Computing},
  volume = {55},
  pages = {100972},
  year = {2025},
  issn = {1875-9521},
  doi = {10.1016/j.entcom.2025.100972},
  author = {van Blerck, Irene C.E. and de Lima, Edirlei Soares and Neggers, Margot M.E. and Calders, Toon},
}

\newpage
\begin{appendices}

\section{Work Identification Prompt}
\label{appendixA}

\subsection{System Prompt}

\textit{You are a narratology expert identifying works that exemplify specific genres through character functions. Focus on structural roles in the story, not character depth or complexity.}

\subsection{Task Prompt}

\textit{Identify a work (novel, play, film, or TV series) in the \{ $G$ \} genre where characters clearly fulfill these four character functions:}

\vspace{0.3cm}

\textit{1. }\{ $R_1$ \}

\textit{2. }\{ $R_2$ \}  

\textit{3. }\{ $R_3$ \}

\textit{4. }\{ $R_4$ \}

\vspace{0.3cm}

\textit{Focus on what characters do in the story structure (their character function), not their psychological complexity.}

\vspace{0.3cm}

\textit{Example format:}

\vspace{0.3cm}

\textit{Title: Le Bourgeois Gentilhomme}

\textit{- lad in love: Cleonte}

\textit{- troubleshooter: Covielle}

\textit{- pompous blocker: Jourdain}

\textit{- witty damsel in love: Lucile}

\vspace{0.3cm}

\textit{Provide your response in this exact format:}

\textit{OUTPUT\_START}

\textit{TITLE: [Title]}

- \{ $R_1$ \}\textit{: [Character name]}

- \{ $R_2$ \}\textit{: [Character name]}

- \{ $R_3$ \}\textit{: [Character name]}

- \{ $R_4$ \}\textit{: [Character name]}

\textit{OUTPUT\_END}

\newpage
\section{Positive Sample Dataset}
\label{appendixB}

Table~\ref{tab:positive_samples} presents the complete set of 40 works with their validated character-role assignments across the four fundamental genres. These correspondences were identified through a semi-automated process combining LLM-assisted generation with expert validation, as described in Section~\ref{subsec:sample_generation}.

\begin{small}
\setlength{\tabcolsep}{3pt}
\begin{longtable}{llllll}
\caption{Complete character-role assignments for the positive sample dataset.}
\label{tab:positive_samples} \\
\toprule
\textbf{Genre} & \textbf{Title} & \textbf{Role 1} & \textbf{Role 2} & \textbf{Role 3} & \textbf{Role 4} \\
\midrule
\endfirsthead
\toprule
\textbf{Genre} & \textbf{Title} & \textbf{Role 1} & \textbf{Role 2} & \textbf{Role 3} & \textbf{Role 4} \\
\midrule
\endhead
\midrule
\multicolumn{6}{r}{\textit{Continued on next page}} \\
\endfoot
\bottomrule
\endlastfoot

\multirow{11}{*}{\rotatebox[origin=c]{90}{\textbf{Comedy}}}
 & & \textit{lad in love} & \textit{troubleshooter} & \textit{pompous blocker} & \textit{witty damsel in love} \\
\cmidrule(l){2-6}
 & Le Bourgeois Gentilhomme & Cleonte & Covielle & Jourdain & Lucile \\
 & Much Ado About Nothing & Claudio & Don Pedro & Leonato & Beatrice \\
 & The Importance of Being Earnest & Jack Worthing & Algernon Moncrieff & Lady Bracknell & Gwendolen Fairfax \\
 & She Stoops to Conquer & Young Marlow & Tony Lumpkin & Mr.\ Hardcastle & Kate Hardcastle \\
 & The Rivals & \makecell[l]{Captain Jack\\Absolute} & Fag & \makecell[l]{Sir Anthony\\Absolute} & Lydia Languish \\
 & A Midsummer Night's Dream & Lysander & Puck & Egeus & Hermia \\
 & The Barber of Seville & Count Almaviva & Figaro & Doctor Bartolo & Rosina \\
 & The School for Scandal & Charles Surface & Sir Oliver Surface & Joseph Surface & Maria \\
 & The Taming of the Shrew & Lucentio & Tranio & Baptista Minola & Bianca \\
 & The Marriage of Figaro & Count Almaviva & Figaro & Doctor Bartolo & Susanna \\
\midrule

\multirow{11}{*}{\rotatebox[origin=c]{90}{\textbf{Romance}}}
 & & \textit{hero} & \textit{donor} & \textit{villain} & \textit{faithful victim} \\
\cmidrule(l){2-6}
 & Ramayana & Rama & Visvamitra & Ravana & Sita \\
 & Pride and Prejudice & Elizabeth Bennet & Mr.\ Darcy & George Wickham & Lydia Bennet \\
 & Jane Eyre & Jane Eyre & Mrs.\ Fairfax & Bertha Mason & Rochester \\
 & Beauty and the Beast (1991) & Belle & Enchantress/Mrs. Potts & Gaston & Beast \\
 & Cinderella (1950) & Prince Charming & Fairy Godmother & Lady Tremaine & Cinderella \\
 & The Princess Bride & Westley & Miracle Max & \makecell[l]{Prince\\Humperdinck} & Buttercup \\
 & Pretty Woman & Edward Lewis & Barney Thompson & Philip Stuckey & Vivian Ward \\
 & Notting Hill & William Thacker & Spike & Anna Scott & William Thacker \\
 & Sabrina (1954) & Linus Larrabee & Baron St. Fontanel & David Larrabee & Sabrina Fairchild \\
 & Wuthering Heights & Edgar Linton & Nelly Dean & Heathcliff & Isabella Linton \\
\midrule

\multirow{11}{*}{\rotatebox[origin=c]{90}{\textbf{Tragedy}}}
 & & \textit{ill-fated adventurer} & \textit{soothsayer} & \textit{order restorer} & \textit{ill-fated partner} \\
\cmidrule(l){2-6}
 & Macbeth & Macbeth & Three Witches & Macduff & Lady Macbeth \\
 & Julius Caesar & Brutus & Soothsayer & Octavius & Cassius \\
 & Oedipus Rex & Oedipus & Tiresias & Creon & Jocasta \\
 & Romeo and Juliet & Romeo & Friar Laurence & Prince Escalus & Juliet \\
 & Antony and Cleopatra & Mark Antony & The Soothsayer & Octavius Caesar & Cleopatra \\
 & Hamlet & Hamlet & \makecell[l]{The Ghost of Hamlet's\\Father} & Fortinbras & Ophelia \\
 & Othello & Othello & Emilia & Lodovico & Desdemona \\
 & King Lear & King Lear & The Fool & Edgar & Cordelia \\
 & A Streetcar Named Desire & Blanche DuBois & Mitch & Stanley Kowalski & Stella Kowalski \\
 & Doctor Faustus & Faustus & Old Man & Good Angel & Mephistopheles \\
\midrule

\multirow{11}{*}{\rotatebox[origin=c]{90}{\textbf{Satire}}}
 & & \textit{nonconformist} & \textit{inquisitor} & \textit{dystopian idol} & \textit{rebellion partner} \\
\cmidrule(l){2-6}
 & 1984 & Winston & O'Brien & Big Brother & Julia \\
 & Brave New World & John the Savage & Mustapha Mond & Henry Foster & Bernard Marx \\
 & Brazil (1985) & Sam Lowry & Jack Lint & Mr.\ Helpmann & \makecell[l]{Archibald ``Harry''\\Tuttle} \\
 & The Handmaid's Tale & Offred/June & Aunt Lydia & Serena Joy & Moira \\
 & V for Vendetta & V & Inspector Finch & Adam Sutler & Evey Hammond \\
 & Fahrenheit 451 & Guy Montag & Captain Beatty & Mildred Montag & Clarisse McClellan \\
 & The Prisoner (1967) & Number Six & Number Two & Number One & Nadia \\
 & The Hunger Games & Katniss Everdeen & Effie Trinket & President Snow & Peeta Mellark \\
 & Catch-22 & Yossarian & Clevinger & Colonel Cathcart & Orr \\
 & A Clockwork Orange & Alex DeLarge & Dr.\ Brodsky & \makecell[l]{Minister of the\\Interior} & Pete \\

\end{longtable}
\end{small}

\newpage
\section{Negative Sample Generation Prompt}
\label{appendixC}

\subsection{System Prompt}

\textit{You are a narratology expert creating systematically incorrect character-function mappings for research validation. Focus on creating plausible-but-wrong associations that test analytical discrimination.}

\subsection{Task Prompt}

\textit{You are helping create test cases for validating character function analysis in narratology research.}

\textit{Given this CORRECT character-role mapping for} \{ $T$ \} \textit{(}\{ $G$ \} \textit{genre):}

\vspace{0.3cm}

- \{ $R_1$ \}: \{ $CH_1$ \}

- \{ $R_2$ \}: \{ $CH_2$ \}

- \{ $R_3$ \}: \{ $CH_3$ \}

- \{ $R_4$ \}: \{ $CH_4$ \}

\vspace{0.3cm}

\textit{Generate ONE INCORRECT mapping by applying this error type: \{ $E$ \}}

\vspace{0.3cm}

\textit{ERROR TYPE DEFINITIONS:}

\textit{- role\_swap: Swap two characters between their roles (maintaining same characters, wrong functions)}

\textit{- minor\_character: Replace one major character with a minor/insignificant character from the work}

\vspace{0.3cm}

\textit{IMPORTANT: }

\vspace{0.3cm}

\textit{- Use actual character names from the work}

\textit{- Make it plausible enough that it requires analysis to detect the error}

\textit{- Only change what is needed for the specified error type}

\vspace{0.3cm}

\textit{Provide your response in this exact format:}

\vspace{0.3cm}

\textit{NEGATIVE\_CASE\_START}

\textit{ERROR\_TYPE: {error\_type}}

\textit{- [role name]: [character name]}

\textit{- [role name]: [character name]}

\textit{- [role name]: [character name]}

\textit{- [role name]: [character name]}

\textit{EXPLANATION: [brief 1-sentence explanation of why this is incorrect]}

\textit{NEGATIVE\_CASE\_END}

\newpage
\section{Negative Sample Dataset}
\label{appendixD}

Table~\ref{tab:negative_samples} presents the complete set of 20 negative samples across the four fundamental genres with their incorrect character-role assignments highlighted in red and bold. These invalid correspondences were generated through a semi-automated error introduction process with expert validation to ensure plausibility, as described in Section~\ref{subsec:negative_construction}.

\begin{small}
\setlength{\tabcolsep}{3pt}
\begin{longtable}{llllll}
\caption{Complete character-role assignments for the negative sample dataset. Incorrect character-role assignments are highlighted in red and bold.}
\label{tab:negative_samples} \\
\toprule
\textbf{Genre} & \textbf{Title} & \textbf{Role 1} & \textbf{Role 2} & \textbf{Role 3} & \textbf{Role 4} \\
\midrule
\endfirsthead
\toprule
\textbf{Genre} & \textbf{Title} & \textbf{Role 1} & \textbf{Role 2} & \textbf{Role 3} & \textbf{Role 4} \\
\midrule
\endhead
\midrule
\multicolumn{6}{r}{\textit{Continued on next page}} \\
\endfoot
\bottomrule
\endlastfoot

\multirow{6}{*}{\rotatebox[origin=c]{90}{\textbf{Comedy}}}
 & & \textit{lad in love} & \textit{troubleshooter} & \textit{pompous blocker} & \textit{witty damsel in love} \\
\cmidrule(l){2-6}
& The Importance of Being Earnest & Jack Worthing & Algernon Moncrieff & Lady Bracknell & \textbf{\textcolor{red}{Miss Prism}} \\
& Le Bourgeois Gentilhomme & \textbf{\textcolor{red}{Covielle}} & \textbf{\textcolor{red}{Cleonte}} & Jourdain & Lucile \\
& The School for Scandal & Charles Surface & Sir Oliver Surface & \textbf{\textcolor{red}{\makecell[l]{Sir Benjamin\\Backbite}}} & Maria \\
& The Taming of the Shrew & \textbf{\textcolor{red}{Tranio}} & \textbf{\textcolor{red}{Lucentio}} & Baptista Minola & Bianca \\
& The Rivals & Captain Jack Absolute & Fag & \makecell[l]{Sir Anthony\\Absolute} & \textbf{\textcolor{red}{Lucy}} \\
\midrule

\multirow{6}{*}{\rotatebox[origin=c]{90}{\textbf{Romance}}}
 & & \textit{hero} & \textit{donor} & \textit{villain} & \textit{faithful victim} \\
\cmidrule(l){2-6}
& Pride and Prejudice & Elizabeth Bennet & \textbf{\textcolor{red}{George Wickham}} & \textbf{\textcolor{red}{Mr. Darcy}} & Lydia Bennet \\
& Cinderella (1950) & \textbf{\textcolor{red}{Fairy Godmother}} & \textbf{\textcolor{red}{Prince Charming}} & Lady Tremaine & Cinderella \\
& Beauty and the Beast (1991) & Belle & \makecell[l]{Enchantress/Mrs.\\Potts} & Gaston & \textbf{\textcolor{red}{Maurice}} \\
& The Princess Bride & Westley & Miracle Max & \textbf{\textcolor{red}{Buttercup}} & \textbf{\textcolor{red}{\makecell[l]{Prince\\Humperdinck}}} \\
& Ramayana & Rama & Visvamitra & Ravana & \textbf{\textcolor{red}{Urmila}} \\
\midrule

\multirow{6}{*}{\rotatebox[origin=c]{90}{\textbf{Tragedy}}}
 & & \textit{ill-fated adventurer} & \textit{soothsayer} & \textit{order restorer} & \textit{ill-fated partner} \\
\cmidrule(l){2-6}
& Antony and Cleopatra & Mark Antony & The Soothsayer & \textbf{\textcolor{red}{Cleopatra}} & \textbf{\textcolor{red}{Octavius Caesar}} \\
& Macbeth & Macbeth & three witches & Macduff & \textbf{\textcolor{red}{Fleance}} \\
& Hamlet & Hamlet & \makecell[l]{The Ghost of Hamlet's\\Father} & Fortinbras & \textbf{\textcolor{red}{Rosencrantz}} \\
& Romeo and Juliet & \textbf{\textcolor{red}{Balthasar}} & Friar Laurence & Prince Escalus & Juliet \\
& A Streetcar Named Desire & Blanche DuBois & \textbf{\textcolor{red}{Stanley Kowalski}} & \textbf{\textcolor{red}{Mitch}} & Stella Kowalski \\
\midrule

\multirow{6}{*}{\rotatebox[origin=c]{90}{\textbf{Satire}}}
 & & \textit{nonconformist} & \textit{inquisitor} & \textit{dystopian idol} & \textit{rebellion partner} \\
\cmidrule(l){2-6}
& The Handmaid's Tale & Offred/June & Aunt Lydia & \textbf{\textcolor{red}{Mrs. Putnam}} & Moira \\
& Fahrenheit 451 & Guy Montag & \textbf{\textcolor{red}{Clarisse McClellan}} & Mildred Montag & \textbf{\textcolor{red}{Captain Beatty}} \\
& The Prisoner (1967) & \textbf{\textcolor{red}{Number Two}} & \textbf{\textcolor{red}{Number Six}} & Number One & Nadia \\
& A Clockwork Orange & \textbf{\textcolor{red}{Dr. Brodsky}} & \textbf{\textcolor{red}{Alex DeLarge}} & \makecell[l]{Minister of the\\Interior} & Pete \\
& Brave New World & John the Savage & Mustapha Mond & Henry Foster & \textbf{\textcolor{red}{\makecell[l]{Helmholtz\\Watson}}} \\
\end{longtable}
\end{small}

\newpage
\section{Validation Prompt}
\label{appendixE}

\subsection{System Prompt}

\textit{You are a narratology expert evaluating character-role correspondences based on character function analysis.}

\textit{Base your analysis on structural narratology frameworks and focus on plot-level functions rather than thematic or psychological interpretations.}

\textit{When multiple characters perform similar functions, identify which one is the primary bearer of that role (the character who most centrally drives that function in the plot structure).}

\subsection{Task Prompt}

\textit{Analyze the following character-role correspondences for their character function justification:}

\vspace{0.3cm}

\textit{Title:} \{ $T$ \}

\textit{Genre:} \{ $G$ \}

\textit{Character-Role Mappings:}

- \{ $R_1$ \}: \{ $CH_1$ \}

- \{ $R_2$ \}: \{ $CH_2$ \}

- \{ $R_3$ \}: \{ $CH_3$ \}

- \{ $R_4$ \}: \{ $CH_4$ \}

\vspace{0.3cm}

\textit{For each character-role correspondence listed above, evaluate whether the character is the primary 
bearer of that role function in the narrative. A correspondence is justified only if this character 
is the main or central character performing that specific function, not a secondary or supporting 
character who also performs related actions.}

\vspace{0.3cm}

\textit{Provide your analysis in the following structured format:}

\vspace{0.3cm}

\textit{ANALYSIS\_START}

---

\textit{ROLE:} \{ $R_1$ \}

\textit{CHARACTER:} \{ $CH_1$ \}

\textit{JUSTIFIED: [YES or NO]}

\textit{REASONING: [Your explanation in 2-3 sentences]}

---

\textit{ROLE:} \{ $R_2$ \}

\textit{CHARACTER:} \{ $CH_2$ \}

\textit{JUSTIFIED: [YES or NO]}

\textit{REASONING: [Your explanation in 2-3 sentences]}

---

\textit{ROLE:} \{ $R_3$ \}

\textit{CHARACTER:} \{ $CH_3$ \}

\textit{JUSTIFIED: [YES or NO]}

\textit{REASONING: [Your explanation in 2-3 sentences]}

---

\textit{ROLE:} \{ $R_4$ \}

\textit{CHARACTER:} \{ $CH_4$ \}

\textit{JUSTIFIED: [YES or NO]}

\textit{REASONING: [Your explanation in 2-3 sentences]}

---

\textit{ANALYSIS\_END}

\newpage
\section{Pairwise Model Agreement}
\label{appendixF}

Table~\ref{tab:pairwise_agreement_detailed} presents the complete pairwise agreement rates between all model pairs across 190 character-role evaluations.

\begin{table}[htbp]
\centering
\caption{Detailed pairwise agreement rates between all model pairs.}
\label{tab:pairwise_agreement_detailed}
\small
\begin{tabular}{llcc}
\toprule
\textbf{Model 1} & \textbf{Model 2} & \textbf{Agreements} & \textbf{Agreement (\%)} \\
\midrule
Claude Opus 4.5 & Llama 4 (128B-A17B) & 166 & 87.4 \\
Claude Opus 4.5 & Gemini 2.5 Pro & 165 & 86.8 \\
Claude Opus 4.5 & Qwen 3 (235B) & 165 & 86.8 \\
Llama 4 (128B-A17B) & Qwen 3 (235B) & 161 & 84.7 \\
Gemini 2.5 Pro & Qwen 3 (235B) & 158 & 83.2 \\
Gemini 2.5 Pro & Llama 4 (128B-A17B) & 157 & 82.6 \\
GPT OSS (120B) & Qwen 3 (235B) & 157 & 82.6 \\
DeepSeek R1 (671B) & Gemini 2.5 Pro & 156 & 82.1 \\
DeepSeek R1 (671B) & Llama 4 (128B-A17B) & 155 & 81.6 \\
DeepSeek R1 (671B) & Qwen 3 (235B) & 154 & 81.1 \\
DeepSeek R1 (671B) & GPT OSS (120B) & 151 & 79.5 \\
Claude Opus 4.5 & GPT OSS (120B) & 150 & 78.9 \\
Claude Opus 4.5 & DeepSeek R1 (671B) & 149 & 78.4 \\
GPT OSS (120B) & Llama 4 (128B-A17B) & 148 & 77.9 \\
Gemini 2.5 Pro & GPT OSS (120B) & 145 & 76.3 \\
\bottomrule
\end{tabular}
\end{table}

\end{appendices} 

\end{document}